\documentclass[review]{elsarticle}
\usepackage{hyperref}
\usepackage{amsmath}
\usepackage{amsthm}
\usepackage{amsfonts}
\usepackage{amssymb}
\usepackage{mathrsfs}
\usepackage{picture}
\usepackage{graphicx}
\usepackage{times}
\usepackage{balance}
\usepackage{booktabs}
\usepackage{multirow}
\usepackage{rotating}
\usepackage{comment}
\usepackage{array}
\usepackage{makecell}
\usepackage{indentfirst}
\usepackage{multirow}  
\usepackage{multicol}
\usepackage{color}
\usepackage{url}
\usepackage{epsfig}
\usepackage{tikz}
\usetikzlibrary{arrows,shapes.geometric,chains}
\usepackage{palatino}
\usepackage{xtab}
\newcommand{\tabincell}[2]{\begin{tabular}{@{}#1@{}}#2\end{tabular}}

\journal{Pattern Recognition}

\begin{document}

\begin{frontmatter}

\title{Multi-modal Visual Tracking: Review and Experimental Comparison}

\author{Pengyu Zhang, Dong Wang, Huchuan Lu}

\begin{abstract}
Visual object tracking, as a fundamental task in computer vision, has drawn much attention in recent years. 
To extend trackers to a wider range of applications, researchers have introduced information from multiple modalities to handle specific scenes, which is a promising research prospect with emerging methods and benchmarks. 
To provide a thorough review of multi-modal tracking, we summarize the multi-modal tracking algorithms, especially visible-depth~(RGB-D) tracking and visible-thermal~(RGB-T) tracking in a unified taxonomy from different aspects. 
Second, we provide a detailed description of the related benchmarks and challenges. 
Furthermore, we conduct extensive experiments to analyze the effectiveness of trackers 
on five datasets: PTB, VOT19-RGBD, GTOT, RGBT234, and VOT19-RGBT.
Finally, we discuss various future directions from different perspectives, including model design and dataset construction for further research. 
\end{abstract}


\end{frontmatter}

\section{Introduction}
Visual object tracking is a fundamental task in computer vision, which is widely applied in many areas, such as smart surveillance, autonomous driving, and human-computer interaction.
Traditional tracking methods are mainly based on visible~(RGB) images captured by a monocular camera. 
When the target suffers long-term occlusion or is in low-illumination scenes, the RGB tracker can hardly work well and may cause tracking failure. 
With the easy-access binocular camera, tracking with multi-modal information (e.g. visible-depth, visible-thermal, visible-radar and visible-laser) is a prospective research direction that has become popular in recent years. 
Many datasets and challenges have been presented~\cite{Li_TIP16_GTOT,Li_MM17_RGBT210,Li_PR19_RGBT234,Xiao_TC18_ARIDM,Kristan_ECCVW20_VOT20,Kristan_ICCVW19_VOT19}. 
Motivated by these developments, trackers with multi-modal cues have been proposed 
with the potential accuracy and robustness against extreme tracking scenarios~\cite{Kart_CVPR19_VSDCF,Cvejic_CVPR07_PLF,Kart_ECCVW18_CSRDCF-rgbd,Zhang_arxiv20_JMMAC,Li_ICCVW19_MANet}. 

With the emergence of multi-modal trackers, a comprehensive and in-depth survey has 
not been conducted. 
To this end, we revisit existing methods from a unified view and evaluate them on popular datasets.
The contributions of this work can be summarized as follows. 
\begin{itemize}
	\item {\bf Substantial review of multi-modal tracking methods from various aspects in a unified view.} We exploit the similarity of RGB-D and RGB-T tracking and classify them in a unified framework. 
	We category existing 56 multi-modal tracking methods based on auxiliary modality,
	tracking framework, and related datasets with corresponding metrics. 
	Taxonomy with detailed analysis can cover the main knowledge in this field, 
	and provide an in-depth introduction to multi-modal tracking models.
	\item {\bf A comprehensive and fair evaluation of popular trackers on several datasets.} We collect 29 methods consisting of 14 RGB-D and 15 RGB-T trackers, and evaluate them 
	on 5 datasets in accuracy and speed for various applications. 
	We further analyze the advantages and drawbacks of different frameworks in qualitative and quantitative experiments.
	\item {\bf A prospective discussion for multi-modal tracking.} We present the potential direction of multi-modal tracking in model design and dataset construction, which can provide prospective guidance to researchers.
\end{itemize}

The rest of the paper is organized as follows.
In section~\ref{Sec:Related_work}, we introduce existing related basic concepts and previous related surveys. 
Section~\ref{Sec:Method} provides a taxonomical review of multi-modal tracking.
We represent the introduction of existing datasets, challenges, and corresponding evaluation metrics described in section~\ref{Sec:Dataset}.
In section~\ref{Sec:Experiment}, we report the experimental results on several datasets and different challenges.
Finally, we discuss the future direction of multi-modal tracking in section~\ref{Sec:Discussion}.
All the collected materials and analysis will be released at \url{https://github.com/zhang-pengyu/Multimodal_tracking_survey}.
\vspace{-5mm}
\section{Background}\label{Sec:Related_work}
\subsection{Visual Object Tracking}
Visual object tracking aims to estimate the coordinates and scales of a specific target throughout the given video. 
In general, tracking methods can be divided into two types according to used information: 
(1) single-modal tracking and (2) multi-modal tracking. Single-modal tracking locates the target captured by a single sensor, such as laser, visible and infrared cameras, to name a few.
In previous years, tracking with RGB image, being computationally efficient, easily 
accessible and high-quality, became increasingly popular. Numerous methods have 
been proposed to improve tracking accuracy and speed.

In RGB tracking, several frameworks including Kalman filter~(KF)~\cite{Weng_JVCIR06_AKF,Kulikov_TSP16_ACDE}, particle filter~(PF)~\cite{YDD_ICCV05_HPF,Okuma_ECCV04_BPF}, sparse learning~(SL)~\cite{Zhang_ECCV12_LRSL,Zhang_CVPR12_MTSL}, correlation filter~(CF)~\cite{Bolme_CVPR10_MOSSE,Li_ECCVW14_SAMF}, and CNN~\cite{Bertinetto_ECCVW16_SiamFC,Zhang_CVPR19_SiamDW} have been involved 
to improve the tracking accuracy and speed.
In 2010, Bolme \emph{et al.}~\cite{Bolme_CVPR10_MOSSE} proposed a CF-based method called 
MOSSE, which achieves high-speed tracking with reasonable performance. 
Thereafter, many researchers have aimed to develop the CF framework to achieve state-of-the-art performance. 
Li \emph{et al.}~\cite{Li_ECCVW14_SAMF} achieve scale estimation and multiple feature integration on the CF framework. 
Martin \emph{et al.}~\cite{Martin_ICCV15_SRDCF} eliminate the boundary effect by adding a spatial regularization to the learned filter at the cost of speed decrease. 
Galoogahi \emph{et al.}~\cite{Galoogahi_ICCV17_BACF} provide another efficient solution to solve the boundary effect, thereby maintaining a real-time speed.
Another popular framework is Siamese-based network, which is first introduced 
by Bertinetto \emph{et al.}~\cite{Bertinetto_ECCVW16_SiamFC}. 
Then, deeper and wider networks are utilized to improve target representation. 
Zhang \emph{et al.}~\cite{Zhang_CVPR19_SiamDW} find that the padding operation 
in the deeper network induces a position bias, suppressing the capability of powerful network. 
They address the position bias problem, and improve the tracking performance significantly. 
Some methods perform better scale estimation by predicting segmentation masks rather than bounding boxes~\cite{Wang_CVPR19_SiamMask,Lukezic_CVPR2020_D3S}.
Above all, many efforts have been conducted in this field. 
However, target appearance, as the main cue from visible images, is not reliable for tracking when target suffers extreme scenarios including low illumination, out-of-view and heavy occlusion. 
To this end, more complementary cues are added to handle these challenges. 
A visible camera is assisted by other sensors, such as laser~\cite{Song_TIST13_RGB-laser}, depth~\cite{Kart_CVPR19_VSDCF}, thermal~\cite{Zhang_arxiv20_JMMAC}, radar~\cite{Kim_IF14_RGB-radar}, and audio~\cite{Megherbi_AVSBS05_RGB-audio}, to satisfy different requirements. 

Since 2005, series of methods have been proposed using various multi-modal information.
Song \emph{et al.}~\cite{Song_TIST13_RGB-laser} conduct multiple object tracking by using visible and laser data. 
Kim \emph{et al.}~\cite{Kim_IF14_RGB-radar} exploit the traditional Kalman filter method 
for multiple object tracking with radar and visible images. 
Megherbi \emph{et al.}~\cite{Megherbi_AVSBS05_RGB-audio} propose a tracking method by combining vision and audio information using belief theory. 
In particular, tracking with RGB-D and RGB-T data has been the focus of attention 
using a portable and affordable binocular camera. 
Thermal data can provide a powerful supplement to RGB images in some challenging 
scenes, including night, fog, and rainy. 
Besides, a depth map can provide an additional constrain to avoid tracking failure 
caused by heavy occlusion and model drift.
Lan \emph{et al.}~\cite{Lan_AAAI18_RCDL} apply the sparse learning method to RGB-T 
tracking, thereby removing the cross-modality discrepancy.
Li \emph{et al.}~\cite{Li_ICCVW19_MANet} extend an RGB tracker to the RGB-T domain, 
which achieves promising results.
Zhang \emph{et al.}~\cite{Zhang_arxiv20_JMMAC} jointly model motion and appearance 
information to achieve accurate and robust tracking. 
Kart \emph{et al.}~\cite{Kart_CVPR19_VSDCF} introduce an effective constraint using 
a depth map to guide model learning.
Liu \emph{et al.}~\cite{Liu_TMM19_TAMS} transform the target position to 3D coordinate 
using RGB and depth images, and then perform tracking using the mean shift method.

\begin{table}[t]
	\tiny
	\vspace{-5mm}
	\caption{Summary of existing surveys in related fields.}
	\label{Table_Related_Survey}
	\begin{center}  
		\begin{tabular}{ccccp{4.2cm}<{\centering}c}  
			\toprule
			Index & Year & Reference & Area & Description & Publication\\
			\midrule
			\multirow{2}{*}{1} & \multirow{2}{*}{2010} & \multirow{2}{*}{\cite{Atrey_MS10_MF_survey}} & \multirow{2}{*}{Multi-modal fusion} & This paper provides a overview on how to fuse multimodal data. & \multirow{2}{*}{MS} \\
			\hline
			\multirow{2}{*}{2} & \multirow{2}{*}{2016} & \multirow{2}{*}{\cite{Walia_AIR16_MM_tracking_survey}} & \multirow{2}{*}{multimodal object tracking} & This paper provides a general review of both single-modal and multi-modal tracking methods. & \multirow{2}{*}{AIR} \\
			\hline
			\multirow{3}{*}{3} & \multirow{3}{*}{2016} & \multirow{3}{*}{\cite{Cai_MTA16_RGBD_survey}} & \multirow{3}{*}{RGB-D dataset} & This paper collects popular RGB-D datasets for different applications and provides an anaysis 
			of the popularity and difficulty. & \multirow{3}{*}{MTA} \\  
			\hline 
			\multirow{2}{*}{4} & \multirow{2}{*}{2017} & \multirow{2}{*}{\cite{Camplani_IETCV17_RGBD_MOT_survey}} & \multirow{2}{*}{RGB-D multiple human tracking} & This paper surveys the existing multiple human tracking methods with RGB-D data into two aspects. & \multirow{2}{*}{IET CV}\\
			\hline
			\multirow{2}{*}{5} & \multirow{2}{*}{2019} & \multirow{2}{*}{\cite{Baltrusaitis_PAMI19_MM_ML_survey}} & \multirow{2}{*}{Multimodal machine learning} & A general survey covers how to represent, translate and fuse multimodal data according to various tasks. & \multirow{2}{*}{TPAMI} \\
			\hline
			\multirow{2}{*}{6} & \multirow{2}{*}{2019} & \multirow{2}{*}{\cite{Ma_IF19_RGBT_IF_survey}} & \multirow{2}{*}{RGB-T Image Fusion} & This paper gives a detailed survey on existing methods and applications for RGB-T image fusion. & \multirow{2}{*}{IF} \\
			\hline
			7 & 2020 & \cite{Zhang_IF20_RGBT_survey} & RGB-T object tracking & A survey of the existing RGB-T tracking methods. & IF \\
			\bottomrule
		\end{tabular}  
	\end{center}  
	\vspace{-5mm}
\end{table}	
\vspace{-4mm}
\subsection{Previous Surveys and Reviews}
As shown in Table~\ref{Table_Related_Survey}, existing surveys are introduced, which are related to multi-modal processing, such as, image fusion, object tracking, and multi-modal machine learning. Some of them focus on specific multi-modal information or single tasks.
Cai \emph{et al.}~\cite{Cai_MTA16_RGBD_survey} collect the datasets captured by RGB-D sensors, which are used in many different applications, such as object recognition, scene classification, hand gesture recognition, 3D-simultaneous localization and mapping, and pose estimation.
Camplani \emph{et al.}\cite{Camplani_IETCV17_RGBD_MOT_survey} focus on multiple human tracking with RGB-D data and conduct an in-depth review of different aspects.
A comprehensive and detailed survey by Ma \emph{et al.}~\cite{Ma_IF19_RGBT_IF_survey} 
is presented to summarize the methods regarding RGB-T image fusion.
Recently, a survey on RGB-T object tracking~\cite{Zhang_IF20_RGBT_survey} is presented, 
which analyzes various RGB-T trackers and conducts quantitative analysis on several 
datasets.

Other surveys aim to give a general introduction on how to utilize and represent 
multi-modal information among a series of tasks.
Atrey \emph{et al.}~\cite{Atrey_MS10_MF_survey} present a brief introduction on 
multi-modal fusion methods and analyze different fusion types in 2010.
Walia \emph{et al.}~\cite{Walia_AIR16_MM_tracking_survey} introduce a general 
survey on tracking with multiple modality data in 2016.
Baltrusaitis \emph{et al.}~\cite{Baltrusaitis_PAMI19_MM_ML_survey} provide a 
detailed review of the machine learning method using multi-modal information.

Various differences and developments are observed among the most related works~\cite{Walia_AIR16_MM_tracking_survey,Zhang_IF20_RGBT_survey}.
First, we aim to conduct a general survey on how to utilize multi-modal 
information, especially RGB-D and RGB-T tracking, on visual object tracking 
in a unified view. 
Furthermore, different from \cite{Walia_AIR16_MM_tracking_survey}, we pay 
much attention to the recent deep-learning-based methods, 
which have not been proposed in 2016. 
Finally, compared with the literature \cite{Zhang_IF20_RGBT_survey} that only 
focuses on RGB-T tracking, our study provides a more substantial and comprehensive 
survey in a large scope, including RGB-D and RGB-T tracking.
\vspace{-5mm}
\section{Multi-modal Visual Tracking}\label{Sec:Method}
This section provides an overview of multi-modal tracking from three aspects: 
(1) auxiliary modality purpose: how to utilize the information of auxiliary 
modality to improve tracking performance; 
(2) tracking framework: the types of framework that trackers belong to. 
Note that, in this study, we mainly focus on visible-thermal (RGB-T), 
visible-depth (RGB-D) tracking, and we consider visible modality as the main modality 
and other sources (i.e. thermal and depth) as auxiliary modalities. 
The taxonomic structure is shown in Figure~1.
\begin{figure*}[t]
	\centering
	\label{fig:taxonomy}
	\caption{Structure of three classification methods and algorithms in each category.}
	\tikzstyle{rec1} = [rectangle, minimum width = 0.54cm, minimum height = 0.3cm, text centered, draw = black]
	\tikzstyle{rec2} = [rectangle, minimum width = 3.5cm, minimum height = 0.3cm, text centered, draw = black]
	\tikzstyle{rec3} = [rectangle, minimum width = 2cm, minimum height = 0.3cm, text centered, draw = black]
	\tikzstyle{rec4} = [rectangle, minimum width = 8.5cm, minimum height = 0.3cm, text centered, draw = black]
	\tikzstyle{rec5} = [rectangle, minimum width = 5cm, minimum height = 0.3cm, text centered, draw = black]
	\tikzstyle{rec6} = [rectangle, minimum width = 5cm, minimum height = 0.3cm, text centered, draw = black]
	\tikzstyle{rec7} = [rectangle, minimum width = 4cm, minimum height = 0.3cm, text centered, draw = black]
	\tikzstyle{arrow2} = [thick, ->, >= stealth]
	\tikzstyle{line} = [draw, -latex']
	\tikzstyle{arrow} = [thick, -, >= stealth]
	\begin{tikzpicture}[node distance = 4cm, auto]
	\node(rec11)[rec1]{\scriptsize \rotatebox{90}{\large{\bf Multimodal Tracking}}};
	\node(rec21)[rec2,right of=rec11,xshift=-1.5cm, yshift = 4cm]{\small \bf Auxiliary Modality Purpose};
	\node(rec22)[rec2,right of=rec11, xshift=-1.5cm, yshift = 0cm]{\small \bf Tracking Framework};
	\node(rec24)[rec2,right of=rec11,xshift=-1.5cm, yshift = -3.5cm]{\small \bf Dataset};
	\node(rec31)[rec3,right of=rec21, xshift=-1cm, yshift = 1cm]{\scriptsize Feature Learning};
	\node(rec32)[rec3,right of=rec21, xshift=-1cm, yshift = -0.8cm]{\scriptsize Pre-processing \cite{Bibi_CVPR16_ASR,Gutev_ICST19_IOTR,Liu_TMM19_TAMS,Xie_CIS19_OD,Zhong_NEU15_3DC,Kart_CVPR19_VSDCF}};
	\node(rec33)[rec3,right of=rec21, xshift=-1cm, yshift = -1.5cm]{\scriptsize Post-processing};
	\node(rec34)[rec3,right of=rec22, xshift=-1cm, yshift = 0.6cm]{\scriptsize Generative};
	\node(rec35)[rec3,right of=rec22, xshift=-1cm, yshift = -0.6cm]{\scriptsize Discriminative};
	
	\node(rec36)[rec3,right of=rec24, xshift=-1cm, yshift = 0cm]{\scriptsize Public Datasets};
	\node(rec37)[rec3,right of=rec24, xshift=-1cm, yshift = -0.9cm]{\scriptsize Challenges\cite{Kristan_ECCVW20_VOT20,Kristan_ICCVW19_VOT19}};
	
	\node(rec41)[rec7,right of=rec31, xshift=0.9cm, yshift = -1.2cm]{\scriptsize\tabincell{c}{EF \cite{An_ICPR16_DLS,Camplani_BMVC2015_ASR,Ding_FSKD15_OHR,Garcia_JDOS12_AMC,Hannuna_RTIP19_DSKCF,Kart_ICPR18_DMDCF,Kuai_Sensors19_TACF,Leng_ACCESS18_SEOH,Liu_Sensors20_WCO,Ma_ITEE17_SL,Meshgi_CVIU16_OAPF,Chen_ISCS08_PGM}\\ \cite{Wu_ICIF11_L1-PF,Li_BICS18_LCSCF,Lan_AAAI18_RCDL,Lan_ACCESS19_FTL,Lan_PRL18_MCASR,Lan_TIE19_MCFT,Li_ECCV18_CMR,Li_MM17_RGBT210,Li_MM16_RT-LSR,Li_PR19_RGBT234,Li_TCSVT19_LGMG,Zhang_Sensor20_MaCNet}\\ \cite{Li_TSMCS17_MLSR,Liu_IS12_JSR,Li_TIP16_GTOT,Zhai_Neu19_CMCF,Gao_ICCVW19_DAFNet,Li_ICCVW19_MANet,Zhu_MM19_DAPNet,Li_NEU18_FTSNet,Yang_ICIP19_TODA,Zhang_CISP18_MDNet-RGBT,Zhang_ICCVW19_DIMP-RGBT,Zhu_arxiv18_FANet}}};
	\node(rec42)[rec5,right of=rec31, xshift=0.9cm, yshift = 0cm]{\scriptsize LF\cite{Conaire_ICIF06_CFM,Shi_CAC15_DG,Wang_Neu14_MCBT,Xiao_TC18_ARIDM,Zhang_CCDC18_RTKCF, Conaire_MVA08_TFF,Cvejic_CVPR07_PLF,Luo_IPT19_AWS,Zhang_arxiv20_JMMAC}};
	\node(rec44)[rec5,right of=rec34, xshift=0.9cm, yshift = 0.7cm]{\scriptsize SL \cite{Ma_ITEE17_SL,Lan_PRL18_MCASR,Lan_AAAI18_RCDL,Lan_ACCESS19_FTL,Lan_TIE19_MCFT,Li_MM16_RT-LSR,Li_TCSVT19_LGMG,Li_TSMCS17_MLSR,Liu_IS12_JSR,Li_TIP16_GTOT,Li_MM17_RGBT210}};
	\node(rec411)[rec5,right of=rec34, xshift=0.9cm, yshift = 0cm]{\scriptsize MS\cite{Gutev_ICST19_IOTR,Liu_TMM19_TAMS,Conaire_MVA08_TFF}};
	
	\node(rec47)[rec5,right of=rec34, xshift=0.9cm, yshift = -0.7cm]{\scriptsize Others\cite{Chen_ISCS08_PGM,Conaire_ICIF06_CFM}};
	
	
	\node(rec45)[rec5,right of=rec35, xshift=0.9cm, yshift = -0.9cm] {\scriptsize\tabincell{c}{CF \cite{An_ICPR16_DLS,Camplani_BMVC2015_ASR,Hannuna_RTIP19_DSKCF,Kart_CVPR19_VSDCF,Kart_ICPR18_DMDCF,Kuai_Sensors19_TACF,Leng_ACCESS18_SEOH,Liu_Sensors20_WCO}\\
			\cite{Xiao_TC18_ARIDM,Zhang_CCDC18_RTKCF,Luo_IPT19_AWS,Zhai_ACCESS18_OACPF,Li_BICS18_LCSCF,Li_NEU18_FTSNet,Zhai_Neu19_CMCF,Kart_ECCVW18_CSRDCF-rgbd,Li_GSKI19_CCF,Zhang_arxiv20_JMMAC}}};
	\node(rec46)[rec5,right of=rec35, xshift=0.9cm, yshift = -1.8cm]{\scriptsize DL\cite{Xie_CIS19_OD,Gao_ICCVW19_DAFNet,Li_ICCVW19_MANet,Zhu_MM19_DAPNet,Yang_ICIP19_TODA,Zhang_CISP18_MDNet-RGBT,Zhang_ICCVW19_DIMP-RGBT,Zhu_arxiv18_FANet,Wang_CVPR20_CMPP,Zhang_Sensor20_MaCNet}};
	\node(rec48)[rec5,right of=rec35, xshift=0.9cm, yshift = 0cm]{\scriptsize PF\cite{Bibi_CVPR16_ASR,Garcia_JDOS12_AMC,Ma_ITEE17_SL,Meshgi_CVIU16_OAPF, Wu_ICIF11_L1-PF,Cvejic_CVPR07_PLF,Liu_IS12_JSR}};
	\node(rec412)[rec5,right of=rec35, xshift=0.9cm, yshift = -2.3cm]{\scriptsize Others\cite{Chen_SP2015_ISOD,Shi_CAC15_DG,Wang_Neu14_MCBT,Zhong_NEU15_3DC,Li_ECCV18_CMR,Ding_FSKD15_OHR,Li_SPIC18_RMR}};
	\node(rec49)[rec5,right of=rec33, xshift=0.9cm, yshift = -0.5cm]{\scriptsize OR\cite{Camplani_BMVC2015_ASR,Chen_SP2015_ISOD,Ding_FSKD15_OHR,Hannuna_RTIP19_DSKCF,Kart_ICPR18_DMDCF,Kuai_Sensors19_TACF,Liu_TMM19_TAMS,Ma_ITEE17_SL,Shi_CAC15_DG,Zhai_ACCESS18_OACPF,Zhang_CCDC18_RTKCF,Zhong_NEU15_3DC,Kart_ECCVW18_CSRDCF-rgbd,Li_GSKI19_CCF,Zhang_arxiv20_JMMAC}};
	\node(rec410)[rec5,right of=rec33, xshift=0.9cm, yshift = 0cm]{\scriptsize SE \cite{Camplani_BMVC2015_ASR,Hannuna_RTIP19_DSKCF,Leng_ACCESS18_SEOH,Zhai_ACCESS18_OACPF,Li_GSKI19_CCF}};
	
	\node(rec413)[rec7,right of=rec36, xshift=0.3cm, yshift = 0cm]{\scriptsize RGB-D\cite{Garcia_JDOS12_AMC,Geiger_IJRR13_KITTI,Liu_ICIP13_ClothStore,Song_ICCV13_PTB,Lukezic_ICCV19_CDTB,Xiao_TC18_ARIDM}};
	\node(rec414)[rec7,right of=rec36, xshift=0.3cm, yshift = -0.5cm]{\scriptsize RGB-T\cite{Davis_CVIU07_OTCBVS,Torabi_CVIU12_LITIV,Li_PR19_RGBT234,Li_TIP16_GTOT,Li_MM17_RGBT210}};
	\draw[line](rec11)|-(rec21);
	\draw[line](rec11)|-(rec22);
	\draw[line](rec11)|-(rec24);
	
	\draw[line](rec21)|-(rec31);
	\draw[line](rec21)|-(rec32);
	\draw[line](rec21)|-(rec33);
	
	\draw[line](rec22)|-(rec34);
	\draw[line](rec22)|-(rec35);
	
	\draw[line](rec24)|-(rec36);
	\draw[line](rec24)|-(rec37);
	
	\draw[line](rec31)|-(rec41);
	\draw[line](rec31)--(rec42);
	
	\draw[line](rec33)|-(rec49);
	\draw[line](rec33)--(rec410);
	
	\draw[line](rec34)|-(rec44);
	\draw[line](rec34)|-(rec411);
	\draw[line](rec34)|-(rec47);
	
	\draw[line](rec35)|-(rec45);
	\draw[line](rec35)|-(rec46);
	\draw[line](rec35)|-(rec48);
	\draw[line](rec35)|-(rec412);
	
	\draw[line](rec36)|-(rec413);
	\draw[line](rec36)|-(rec414);
	\end{tikzpicture}
	\vspace{-10mm}
\end{figure*}
\vspace{-3mm}
\subsection{Auxiliary Modality Purpose}
We first discuss the auxiliary modality purpose in multi-modal tracking. 
There are three main categories: (a) feature learning, where the feature representations 
of auxiliary modality image are extracted to help locate the target;
(b) pre-processing, where the information from auxiliary modality is used before the 
target modeling; and 
(c) post-processing, where the information from auxiliary modality aims to improve 
the model or refine the bounding box.

\subsubsection{Feature Learning}
\begin{figure}[t]
	\centering
	\footnotesize
	\includegraphics[width=1.0\linewidth]{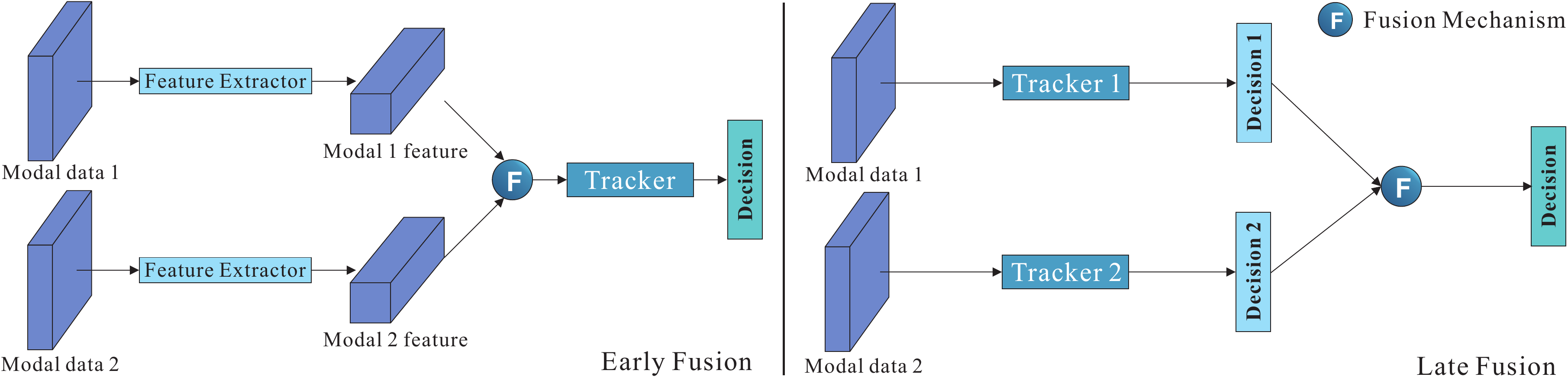}
	\caption{Workflows of early fusion (EF) and late fusion (LF). EF-based methods conduct feature fusion and model them jointly; while LF-based methods aim to model each modality individually and then combine their decisions.}
	\label{fig:fusion}
	\vspace{-15mm}
\end{figure}

Methods based on feature learning extract information from auxiliary modality through
various feature methods, and then adopt modality fusion to combine the data from 
different sources. 
Feature learning is an explicit way to utilize multi-modal information, and most of 
corresponding methods consider the image of auxiliary modality as an extra channel of the model. 
According to different fusion methods, as shown in Figure \ref{fig:fusion}, it can be further categorized as methods based on early fusion (EF) and late fusion (LF)~\cite{Atrey_MS10_MF_survey,Ramachandram_SPM17_DML_survey}. 
EF-based methods combine multi-modal information in the feature level using 
concatenation and summation approaches; while LF-based methods model each modality individually and obtain the final result by considering both decisions of modalities.

{\flushleft \bf Early Fusion (EF)}. In EF-based methods, the features extracted from both modalities are first aggregated as a larger feature vector and then sent to the model to locate the target. 
The workflow of EF-based trackers is shown in the left part of Figure \ref{fig:fusion}. 
For most of the trackers, EF is the primary choice in the multi-modal tracking task, 
while visible and auxiliary modalities are treated alike with the same feature 
extraction methods. 
Camplani \emph{et al.}~\cite{Camplani_BMVC2015_ASR} utilize HOG feature for both 
visible and depth maps. 
Kart \emph{et al.}~\cite{Kart_ICPR18_DMDCF} extract multiple features to build a robust tracker for RGB-D tracking. 
Similar methods exist in~\cite{Ding_FSKD15_OHR,Kuai_Sensors19_TACF,Leng_ACCESS18_SEOH,An_ICPR16_DLS,Wu_ICIF11_L1-PF,Lan_ACCESS19_FTL,Lan_TIE19_MCFT,Li_MM17_RGBT210,Li_MM16_RT-LSR,Li_PR19_RGBT234}. 
However, auxiliary modality often indicates different information against the visible map. 
For example, thermal and depth images contain temperature and depth data, respectively. 
The aforementioned trackers apply feature fusion, ignoring the modality discrepancy, 
which decreases the tracking accuracy and causes the tracker to drift easily.
To this end, some trackers differentiate the heterogeneous modalities by applying 
different feature methods. 
In~\cite{Garcia_JDOS12_AMC}, the gradient feature is extracted in a depth map, 
while the average color feature is used to represent the target in the visible 
modality. 
Meshgi \emph{et al.}~\cite{Meshgi_CVIU16_OAPF} use the raw depth information 
and many feature methods~(HOG, LBP, and LoG) for RGB images. 
In~\cite{Lan_AAAI18_RCDL,Lan_PRL18_MCASR,Liu_IS12_JSR}, the HOG and intensity 
features are used for visible and thermal modalities, respectively.
Due to the increasing cost involved in feature concatenation and the misalignment 
of multi-modal data, some methods tune the feature representation after feature 
extraction by the pruning~\cite{Zhu_MM19_DAPNet} or re-weighting operation~\cite{Liu_Sensors20_WCO,Zhu_arxiv18_FANet}, which can compress the 
feature space and exploit the cross-modal correlation. 
In DAFNet~\cite{Zhu_MM19_DAPNet}, a feature pruning module is proposed to eliminate 
noisy and redundant information. 
Liu \emph{et al.}~\cite{Liu_Sensors20_WCO} introduce a spatial weight to highlight 
the foreground area. 
Zhu et al.~\cite{Zhu_arxiv18_FANet} exploit modality importance using the proposed 
multi-modal aggregation network.

{\flushleft \bf Late fusion (LF)}. LF-based methods process both modalities simultaneously 
and the independent models for each modality are built to make decisions. 
Then, the decisions are combined by using weighted summation~\cite{Luo_IPT19_AWS,Shi_CAC15_DG,Xiao_TC18_ARIDM,Zhang_CCDC18_RTKCF}, calculating joint distribution function~\cite{Conaire_ICIF06_CFM,Cvejic_CVPR07_PLF,Conaire_MVA08_TFF}, 
and conducting multi-step localization~\cite{Wang_Neu14_MCBT}.
Conaire \emph{et al.}~\cite{Conaire_ICIF06_CFM} assume the independence between 
multi-modal data, and then obtain the result by multiplying the target's likelihoods 
in both modalities. 
A similar method is adopted in literature~\cite{Conaire_MVA08_TFF}. 
Xiao \emph{et al.}~\cite{Xiao_TC18_ARIDM} fuse two single-modal trackers via 
an adaptive weight map. 
In MCBT~\cite{Wang_Neu14_MCBT}, data from multiple sources are used stepwise 
to locate the target. 
A rough target position is first estimated by optical flow in the visible domain,  
and the final result is determined by part-based matching method with RGB-D data.
\subsubsection{Pre-Processing}
Due to the available depth map, the second purpose of auxiliary modality is to 
transform the target into 3D space before target modeling via RGB-D data. 
Instead of tracking in the image plane, these types of methods model the target 
in the world coordinate, and 3D trackers are designed~\cite{Bibi_CVPR16_ASR,Gutev_ICST19_IOTR,Kart_CVPR19_VSDCF,Liu_TMM19_TAMS,Xie_CIS19_OD,Zhong_NEU15_3DC}. 
Liu \emph{et al.}~\cite{Liu_TMM19_TAMS} extend the classical mean shift tracker to 3D extension.
In OTR~\cite{Kart_CVPR19_VSDCF}, the dynamic spatial constraint generated by the 3D 
target model enhances the discrimination of DCF trackers in dealing with out-of-view 
rotation and heavy occlusion. 
Although a significant performance is achieved, the computation cost of 3D 
reconstruction cannot be neglected. 
Furthermore, the performance is highly subject to the quality of depth data 
and the accessibility of mapping functions between the 2D and 3D spaces.

\subsubsection{Post-processing}
Compared with the RGB image that brings more detailed content, the depth image 
highlights the contour of objects, which can segment the target among surroundings 
via depth variance. 
Inspired by the nature of depth map, many RGB-D trackers utilize the depth 
information to determine whether the occlusion occurs and estimate the target scale~\cite{Camplani_BMVC2015_ASR,Hannuna_RTIP19_DSKCF,Leng_ACCESS18_SEOH,Zhai_ACCESS18_OACPF}. 

{\flushleft \bf Occlusion Reasoning (OR)}. Occlusion is a traditional challenge in the 
tracking task because the dramatic appearance variation leads the model drifting. 
Depth cue is a powerful feature to detect target occlusion; thus, the tracker 
can apply a global search strategy or model updating mechanism to avoid learning 
from the occluded target. 
In~\cite{Camplani_BMVC2015_ASR}, occlusion is detected when the depth variance 
is large. Then, tracker enlarges the search region to detect the re-appeared target. 
Ding \emph{et al.}~\cite{Ding_FSKD15_OHR} propose an occlusion recovery method, 
where a depth histogram is recorded to examine whether the occlusion occurs. 
If the occlusion is detected, the tracker locates the occluder and searches 
the candidate around. 
In~\cite{Zhang_arxiv20_JMMAC}, Zhang \emph{et al.} propose a tracker switcher 
to detect occlusion based on the template matching method and tracking reliability. 
The tracker can dynamically select which information is used for tracking 
between appearance and motion cues, thereby improving the robustness of the 
tracker significantly.

{\flushleft \bf Scale Estimation (SE)}. SE is an important module in tracking task, 
which can obtain a tight bounding box and avoid drift. 
CF-based trackers estimate the target scale by sampling the search region in 
multiple resolutions~\cite{Li_ECCV14_SAMF}, learning a filter for scale estimation~\cite{Danelljan_PAMI17_DSST}, which cannot effectively adapt to 
the target's scale change~\cite{Leng_ACCESS18_SEOH}.
Both thermal and depth maps provide clear contour information and a coarse 
pixel-wise target segmentation map. 
With such information, the target shape can be effectively estimated.
In \cite{Hannuna_RTIP19_DSKCF}, the number of scales is adaptively 
changed to fit the scale variation. 
SEOH~\cite{Leng_ACCESS18_SEOH} uses space continuity-of-depth information 
to achieve accurate scale estimation with minor time cost. 
The pixels belonging to the target are clustered by the K-means method 
in the depth map, and the sizes of the target and search regions are 
determined by clustering result. 

\subsection{Tracking Framework}
In this section, multi-modal trackers are categorized based on the methods 
used in target modeling, including generative and discriminative. The generative 
framework focuses on directly modeling the representation of the target. 
During tracking, the target is captured by matching the data distribution in the 
incoming frame. However, generative methods only learn the representations for 
the foreground information while ignoring the influence of surroundings, 
suffering from background cluttering or distractions~\cite{Li_TIST13_SOT_survey}.
In comparison, the discriminative models construct an effective classifier to 
distinguish the object against the surroundings. 
The tracker outputs the confidence score of sampled candidates and chooses 
the best matching patch as the target. 
Various patch sample manners are exploited, e.g. sliding window~\cite{Liu_Sensors20_WCO}, particle filter~\cite{Bibi_CVPR16_ASR,Garcia_JDOS12_AMC}, and Gaussian sampling~\cite{Li_ICCVW19_MANet}. 
Furthermore, a crucial task is utilizing powerful features to represent the target. 
Thanks to the emerging convolution networks, more trackers have been built via 
efficient CNNs. We will introduce the various frameworks in the following paragraphs.

\subsubsection{Generative Methods}
\begin{figure}
	\centering
	\vspace{-10mm}
	\includegraphics[width=1.0\linewidth]{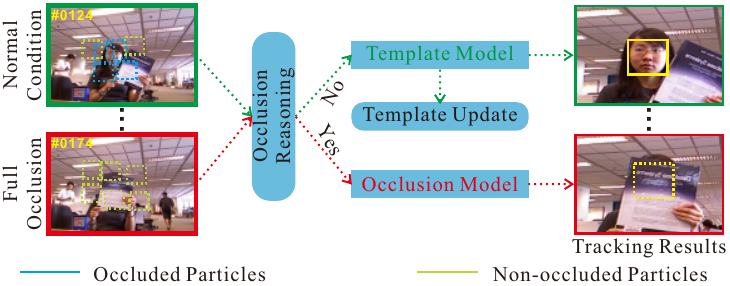}
	\vspace{-10mm}
	\caption{Framework of OAPF~\cite{Meshgi_CVIU16_OAPF}. The particle filter method 
	is applied, with occlusion handling, in which the occlusion model is constructed 
	against the template model. When the target is occluded, the occlusion model is 
	used to predict the position without the updating template model.}
	\label{fig:OAPF}
	\vspace{-3mm}
\end{figure}

{\flushleft \bf Sparse Learning (SL)}. SL has been popular in many tasks including image recognition~\cite{Shojaeilangari_TIP15_SL1} and classification~\cite{Yang_IJCV14_SL2}, 
object tracking~\cite{Xie_TC14_SL3}, and others. 
In SL-based RGB-T trackers, the tracking task can be formulated as a minimization 
problem for the reconstruction error with the learned sparse dictionary~\cite{Lan_PRL18_MCASR,Lan_AAAI18_RCDL,Lan_ACCESS19_FTL,Lan_TIE19_MCFT,Li_MM16_RT-LSR,Li_TSMCS17_MLSR,Liu_IS12_JSR,Li_TIP16_GTOT}. 
Lan \emph{et al.}~\cite{Lan_AAAI18_RCDL} propose a unified learning paradigm to 
learn the target representation, modality-wise reliability and classifier, collaboratively. 
Similar methods are also applied in the RGB-D tracking task. 
Ma \emph{et al.}~\cite{Ma_ITEE17_SL} construct an augmented dictionary consisting 
of target and occlusion templates, which achieves accurate tracking even in heavy occlusion. 
SL-based trackers achieve promising results at the expense of computation cost. 
These trackers cannot meet the requirements of real-time tracking.

{\flushleft \bf Mean Shift (MS)}. MS-based methods maximize the similarity between 
the histograms of candidates and the target template, and conduct fast 
local search using the mean shift technique. These methods usually 
assume that the object overlaps itself in consecutive frames~\cite{Conaire_MVA08_TFF}. 
In~\cite{Gutev_ICST19_IOTR,Liu_TMM19_TAMS}, the authors extend the 2D MS method to 3D 
with RGB-D data. 
Conaire \emph{et al.}~\cite{Conaire_MVA08_TFF} propose an MS tracker using spatiogram 
instead of histogram. 
Compared with discriminative methods, MS-based trackers directly regress the offset 
of the target, which omits dense sampling. 
These methods with lightweight features can achieve real-time performance, whereas 
the performance advantage is not obvious.

{\flushleft \bf Other Frameworks}. Other generative methods have been applied to 
tracking tasks. 
Coraire \emph{et al.}~\cite{Conaire_ICIF06_CFM} model the tracked object via 
Gaussian distribution and select the best-matched patch via similarity measure. 
Chen \emph{et al.}~\cite{Chen_ISCS08_PGM} model the statistics of each individual 
modality and the relationship between RGB and thermal data using the expectation 
maximization algorithm. 
These methods can model individual or complementary modalities, thereby achieving 
a flexible framework for different scenes.

\subsubsection{Discriminative Methods}
\begin{figure}
	\centering
	\includegraphics[width=1.0\linewidth]{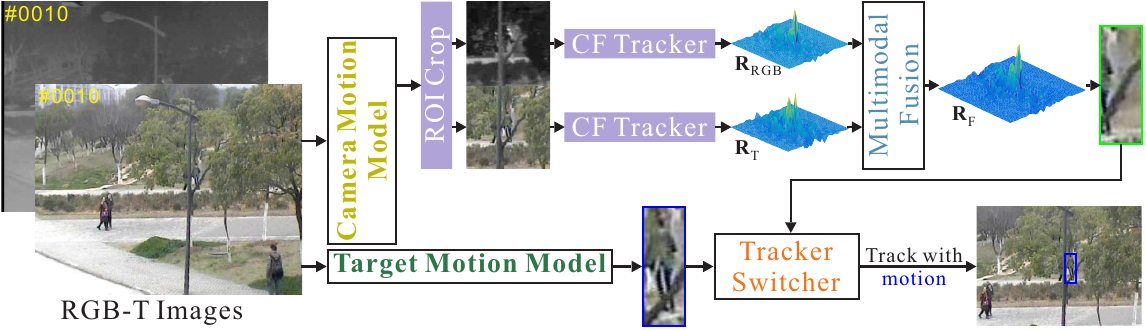}
	\caption{Workflow of the JMMAC~\cite{Zhang_arxiv20_JMMAC}. The CF-based 
	tracker is used to model appearance cue, while both camera and target motion are considered, thereby achieving substantial performance. }
	\label{fig:JMMAC}
	\vspace{-3mm}
\end{figure}

{\flushleft \bf Particle Filter~(PF)}. The PF framework is a Bayesian sequential importance sampling technique~\cite{Isard_IJCV98_PF}. It consists of two steps, i.e., prediction and updating. 
In the prediction step, given the state observations ${\bf z}_{1:t} = \{{\bf z}_1,{\bf z}_2,...,{\bf z}_{t}\}$ during the previous $t$ frames, the posterior distribution of the state $\mathbf{x}_{t}$ is predicted using Bayesian rule as follows: 
\begin{equation}
p\left(\mathbf{x}_{t} \mid \mathbf{z}_{1: t}\right)=\frac{p\left(\mathbf{z}_{t} \mid \mathbf{x}_{t}\right) p\left(\mathbf{x}_{t} \mid \mathbf{z}_{1: t-1}\right)}{p\left(\mathbf{z}_{t} \mid \mathbf{z}_{1: t-1}\right)},
\end{equation}
where $p\left(\mathbf{x}_{t} \mid \mathbf{z}_{1: t-1}\right)$ is estimated by a set of $N$ particles. Each particle has a weight, $w_t^i$.
In the updating process, $w_t^i$ is updated as 
\begin{equation}
w_{t}^{i} \propto p\left({\bf z}_{t} \mid {\bf x}_{t}={\bf x}_{t}^{i}\right).
\end{equation}
In the PF framework, the restrictions of linearity and Gaussianity imposed 
by Kalman filter are relaxed, thereby leading accurate and robust tracking~\cite{Cvejic_CVPR07_PLF}. 
Several works improve the PF method for multi-modal tracking task.
Bibi \emph{et al.}~\cite{Bibi_CVPR16_ASR} formulate the PF framework in 3D, 
which considers both representation and motion models and propose a particle 
pruning method to boost the tracking speed. 
Meshgi \emph{et al.}~\cite{Meshgi_CVIU16_OAPF} consider occlusion in approximation step to improve PF in occlusion handling. 
Liu \emph{et al.}~\cite{Liu_IS12_JSR} propose a new likelihood function for 
PF to determine the goodness of particles, thereby promoting the performance. 

\begin{figure}
	\centering
	\includegraphics[width=1.0\linewidth]{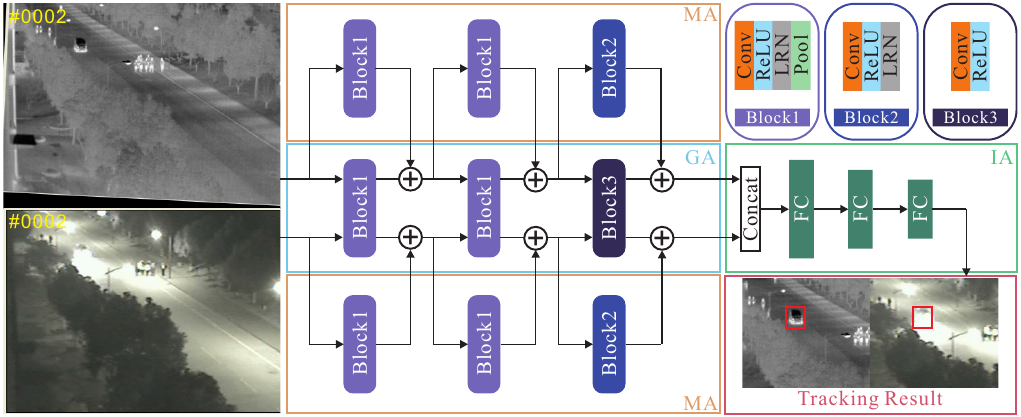}
	\caption{Framework of MANet~\cite{Li_ICCVW19_MANet}. Generic adapter~(GA) is used to extract common information of RGB-T images. Modality adapter~(MA) aims to exploit the different properties of heterogeneous modalities. Finally, instance adapter~(IA) 
	models the appearance properties and temporal variations of a certain object. }
	\label{fig:MANet}
	\vspace{-5mm}
\end{figure}

{\flushleft \bf Correlation Filter~(CF)}. CF-based tracker learns the discriminative 
template denoted as CF to represent the target. 
Then, the online learned filter is used to detect the object in the next frame. 
As the circular convolution can be accelerated in Fourier domain, these trackers 
can maintain approving accuracy with high speed. 
In recent years, many CF-based variants are proposed, such as adding spatial regularization~\cite{Danelljan_ICCV15_SRDCF}, introducing temporal constraint~\cite{Li_CVPR18_STRCF}, and equipping discriminative features~\cite{Danelljan_CVPR17_ECO}, to increase the tracking performance. 
Due to the advantage of CF-based trackers, many researchers aim to build 
multi-modal trackers with the CF framework. 
Zhai~\emph{et al.}~\cite{Zhai_Neu19_CMCF} introduce low-rank constraint to learn 
the filters of both modalities collaboratively, thereby exploiting the relationship 
between RGB and thermal data.
Hannuna~\emph{et al.}~\cite{Hannuna_RTIP19_DSKCF} effectively handle the scale change 
with the guidance of the depth map.
Kart~\emph{et al.} propose a long-term RGB-D tracker~\cite{Kart_CVPR19_VSDCF}, which 
is designed based on CSRDCF~\cite{Lukezic_IJCV18_CSRDCF} and applies online 3D target reconstruction to facilitate learning robust filters. 
The spatial constraint is learned from the 3D model of the target. 
When the target is occluded, view-specific DCFs are used to robustly 
localize the target.
Camplani~\emph{et al.}~\cite{Camplani_BMVC2015_ASR} improve the CF method 
in scale estimation and occlusion handling, while maintaining a real-time speed.

{\flushleft \bf Deep Leraning~(DL)}. Due to the discriminative ability in feature representation, CNN is widely used in the tracking task. 
Various networks provide a powerful alternative to the traditional hand-crafted 
feature, which is the simplest way to utilize CNN. 
Liu \emph{et al.}~\cite{Liu_Sensors20_WCO} extract the deep features from VGGNet~\cite{Simonyan_ICLR15_VGGNet} and hand-crafted features to learn 
a robust representation.
Li \emph{et al.}\cite{Li_NEU18_FTSNet} concatenate deep features from visible 
and thermal images, and then adaptively fuse them using the proposed FusionNet 
to achieve robust feature representation.
Furthermore, some methods aim to learn an end-to-end network for multi-modal tracking.
In~\cite{Li_ICCVW19_MANet,Zhu_MM19_DAPNet,Yang_ICIP19_TODA}, a similar framework 
borrowed from MDNet~\cite{MDNet} is applied for tracking with different structures 
to fuse the cross-modal data. 
These trackers achieve obvious performance promotion while the speed is poor.
Zhang \emph{et al.}~\cite{Zhang_ICCVW19_DIMP-RGBT} propose an end-to-end RGB-T 
tracking framework with real-time speed and balanced accuracy. 
They apply ResNet~\cite{He_CVPR16_ResNet} as the feature extractor and fuse 
RGB and thermal information in the feature level, which are used for target 
localization and box estimation. 

{\flushleft \bf Other Frameworks}.
Some methods use an explicit template matching method to localize the object. 
These methods find the best-matched candidate with the target captured in 
frames through a pre-defined matching function~\cite{Wang_Neu14_MCBT,Zhong_NEU15_3DC}.
Ding \emph{et al.}~\cite{Ding_FSKD15_OHR} learn a Bayesian classifier and 
consider the candidate with maximal score as the target location, 
which can reduce the model drift.
In~\cite{Li_SPIC18_RMR}, a structured SVM~\cite{Hare_ICCV11_Struck} is learned 
by maximizing a classification score, which can prevent the labeling ambiguity in the training process.
\vspace{-5mm}
\section{Datasets}\label{Sec:Dataset}
With the emergence of multi-modal tracking methods, several datasets and challenges 
for RGB-D and RGB-T tracking are released. We summarize the available datasets in Table~\ref{Table_dataset}.

\begin{table}
	\tiny
	\caption{Summary of multi-modal tracking datasets.}
	\label{Table_dataset}
	\begin{center}  
		\begin{tabular}{|c|c|c|c|c|c|c|c|c|c|}  
			\hline
			 & Name & Seq. Num.  & Total Frames & Min. Frame & Max. Frame & Attr. & Resolution & Metrics & Year\\
			\hline
			\multirow{3}{*}{\rotatebox{90}{\tiny{\bf RGB-D}}}& PTB~\cite{Song_ICCV13_PTB} & 100 & 21.5K & 0.04K & 0.90K & 11 & 640 $\times$ 480 & CPE, SR & 2013 \\
			& STC~\cite{Xiao_TC18_ARIDM} & 36 & 18.4K & 0.13K & 0.72K 
			
			& 10 & 640 $\times$ 480 & SR, Acc., Fail. & 2018 \\
			&CTDB~\cite{Lukezic_ICCV19_CDTB} & 80 & 101.9K & 0.4K & 2.5K & 13 & 640 $\times$ 360 & F-score, Pr, Re & 2019 \\
			\hline
			\multirow{6}{*}{\rotatebox{90}{\tiny{\bf RGB-T}}}& OTCBVS~\cite{Davis_CVIU07_OTCBVS}& 6 & 7.2K & 0.6K & 2.3K & -- & 320 $\times$ 240 & -- & 2007\\
			& LITIV~\cite{Torabi_CVIU12_LITIV}& 9 & 6.3K & 0.3K & 1.2K & -- & 320 $\times$ 240 & -- & 2012\\
			& GTOT~\cite{Li_TIP16_GTOT} & 50 & 7.8K & 0.04K & 0.37K & 7 &384 $\times$ 288 & SR, PR & 2016\\
			& RGBT210~\cite{Li_MM17_RGBT210} & 210 & 104.7K & 0.04K & 4.1K & 12 & 630 $\times$ 460 & SR, PR & 2017\\
			& RGBT234~\cite{Li_PR19_RGBT234} & 234 & 116.6K & 0.04K & 8.1K & 12 & 630 $\times$ 460 & SR, PR, EAO & 2019\\
			& VOT-RGBT~\cite{Kristan_ECCVW20_VOT20,Kristan_ICCVW19_VOT19} & 60 & 20.0K & 0.04K & 1.3K & 5 & 630 $\times$ 460 & EAO & 2019 \\
			\hline
		\end{tabular}  
	\end{center}  
	\vspace{-5mm}
\end{table}			
\subsection{Public dataset}

\subsubsection{RGB-D dataset}  
In 2012, a small-scale dataset called BoBoT-D~\cite{Garcia_JDOS12_AMC} is constructed, consisting of five RGB-D video sequences captured by Kinect V1 sensor. 
Both overlap and hit rate are used for evaluation, which indicate the mean overlap between result and ground truth and percentage of frame where the overlap is larger than 0.33.
Song \emph{et al.}~\cite{Song_ICCV13_PTB} propose the well-known Princeton tracking benchmark~(PTB) of 100 high-diversity RGB-D videos, five of which are used for 
validation and others without available ground truth are used for testing. 
The PTB dataset contains 11 annotations, which are separated by 5 categories 
including target type, target size, movement, occlusion, and motion type. 
Two metrics are conducted to evaluate the tracking performance: center position 
error~(CPE) and success rate~(SR). 
CPE measures the Euclidean distance between centers of result and ground truth and 
SR is the average intersection over union~(IoU) during all frames, which is defined 
as  
\begin{equation}\label{SR}
SR = \frac{1}{N} \sum_{i=1}^{N} u_{i} \ \ \ \ \ \ u_{i} = \left\{\begin{matrix}
1 & IoU(bb_{i},gt_i)> t_{sr}\\ 
0 & otherwise
\end{matrix}\right. ,
\end{equation}
where the $IoU(\cdot,\cdot)$ denotes the IoU between the bounding box $bb_i$ 
and ground truth $gt_i$ in the $i$-th frame. 
If the IoU is larger than the threshold $t_{sr}$, we consider the target 
to be successfully tracked.
The final rank of the tracker is determined by the Avg. Rank, which is 
defined as the average ranking of SR in each attribute.
The STC dataset~\cite{Xiao_TC18_ARIDM} consists of 36 RGB-D sequences and 
covers some extreme tracking circumstances, such as outdoor and night scenes. 
This dataset is captured by still and moving ASUS Xtion RGB-D cameras to 
evaluate the tracking performance under conditions of arbitrary camera motion. 
A total of 10 attributes are labeled to thoroughly analyze the dataset bias. \emph{The detailed introduction of each attributes are shown in the supplementary file.}

\begin{figure}
	\centering
	\includegraphics[width=1.0\linewidth]{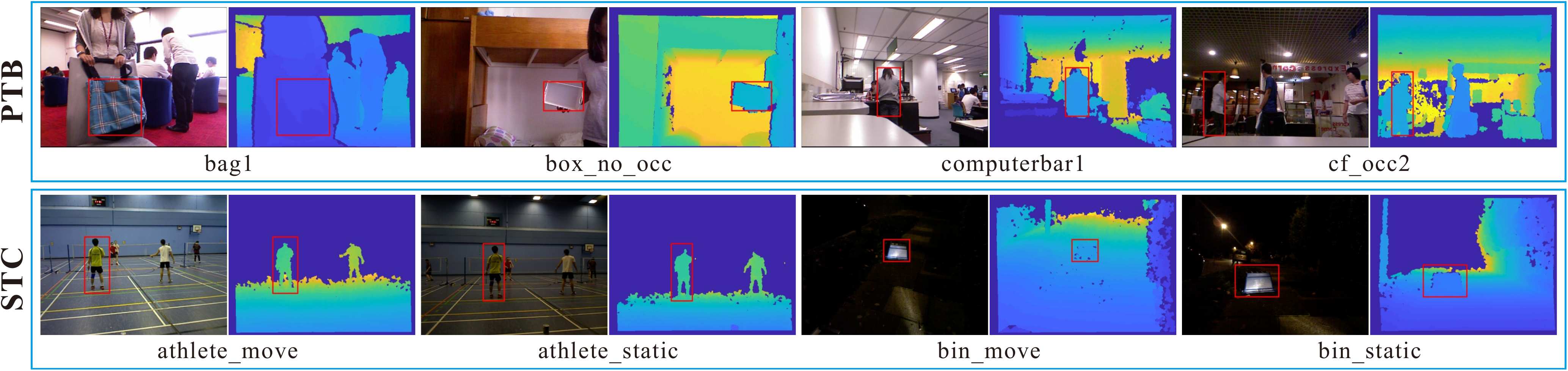}
	\vspace{-10mm}
	\caption{Examples in RGB-D tracking datasets (PTB and STC datasets).}
	\label{fig:RGBD_example}
	\vspace{-5mm}
\end{figure}

The trackers are measured by using both SR and VOT protocols. 
The VOT protocol evaluates the tracking performance in terms of two 
aspects: accuracy and failure.
Accuracy~(Acc.) considers the IoU between the ground truth and bounding box, 
and failure~(Fail.) measures the times when the overlap is zero and the tracker 
is set to re-initialize using the ground truth and continues to track.
CTDB~\cite{Lukezic_ICCV19_CDTB} is the latest RGB-D tracking dataset, which 
contains 80 short-term and long-term videos. 
The target is out-of-view and occluded frequently, which needs the tracker to 
handle both tracking and re-detection cases. 
The metrics are Precision~(Pr.), Recall~(Re.) and the overall F-score~\cite{Lukezic_arxiv18_LT_metric}. 
The precision and recall are defined as follows, 

\begin{equation}
 Pr = \frac{\sum_{i=1}^{N}u_i}{\sum_{i=1}^{N}s_i} \ \ \ u_{t} = \left\{\begin{matrix}
1 & bb_i  \ \ exists\\ 
0 & otherwise
\end{matrix}\right. ,
\end{equation}

\begin{equation}
Re = \frac{\sum_{i=1}^{N}u_i}{\sum_{i=1}^{N}g_i} \ \ \ g_{t} = \left\{\begin{matrix}
1 & gt_i  \ \ exists\\ 
0 & otherwise
\end{matrix}\right. , 
\end{equation}
where $u_i$ is defined in Eq.~\ref{SR}. 
The F-score combines both precision and recall through  
\begin{equation}
F-score = \frac{2Pr \times Re}{Pr + Re}. 
\end{equation}

\vspace{-2mm}
\subsubsection{RGB-T Dataset} In previous years, two RGB-T people detection 
datasets are used for tracking. The OTCBVS dataset~\cite{Davis_CVIU07_OTCBVS} 
has six grayscale-thermal video clips captured from two outdoor scenes. 
The LITIV dataset~\cite{Torabi_CVIU12_LITIV} contains nine sequences, 
considering the illumination influence and being captured indoors. 
These datasets with limited sequences and low diversity have been depreciated.
In 2016, Li \emph{et al.} construct the GTOT dataset for RGB-T tracking, which 
consists of 50 grayscale-thermal sequences under different scenarios and conditions. 
A new attribute for RGB-T tracking is labeled as thermal crossover~(TC), which 
indicates that the target has similar temperature with the background. 
Inspired by~\cite{Wu_CVPR13_OTB13,Wu_PAMI15_OTB15}, GTOT adopts success rate~(SR) and precision rate~(PR) for evaluation. PR denotes the percentage of frames whose CPE is 
smaller than a threshold $t_{pr}$, which is set to 5 in GTOT to evaluate small targets.
Li \emph{et al.}~\cite{Li_MM17_RGBT210} propose a large-scale RGB-T tracking dataset, 
namely RGBT210, which contains 210 videos and 104.7k image pairs. 
This dataset also extends the number of attributes to 12. \emph{The detailed description of attributes can be found in supplementary file.} The metric is the same as GTOT, except $t_{pr}$ is set to 20 normally.
In 2019, the researchers enlarge the RGBT210 dataset and propose RGBT234~\cite{Li_PR19_RGBT234}, which provides individual ground truth for each modality. Furthermore, besides SR and PR, expected average overlap~(EAO) is used for evaluation, combining the accuracy and failures in a principled manner. 

\begin{figure}
	\centering
	\includegraphics[width=1.0\linewidth]{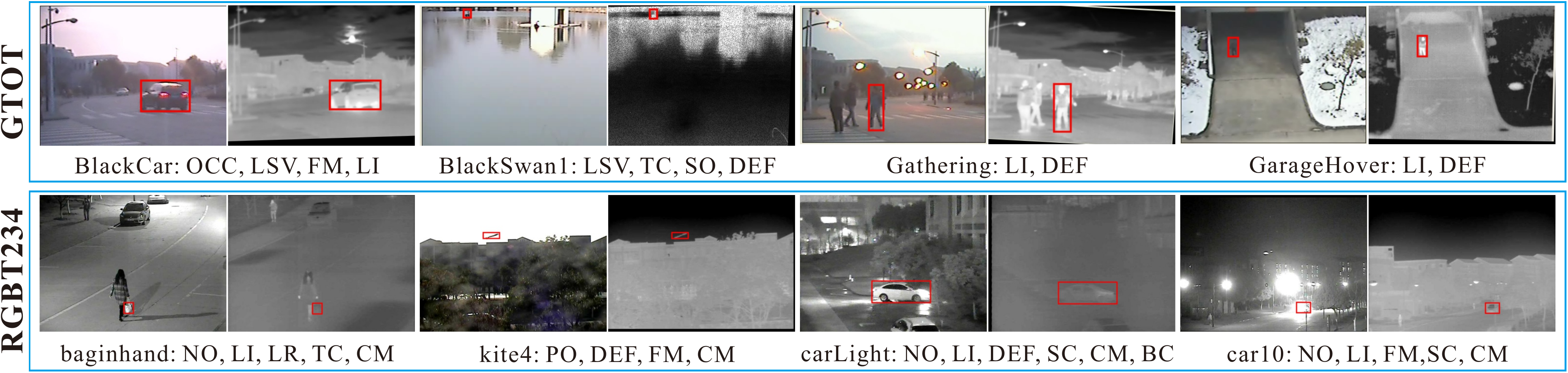}
	\vspace{-10mm}
	\caption{Examples and corresponding attributes in GTOT and RGBT234 tracking datasets.}
	\vspace{-5mm}
	\label{fig:RGBT_example}
\end{figure}

\subsection{Challenges for Multi-modal Tracking}
Since 2019, both RGB-D and RGB-T challenges have been held by VOT Committee~\cite{Kristan_ICCVW19_VOT19,Kristan_ECCVW20_VOT20}. 
For RGB-D challenge, trackers are evaluated on the CDTB dataset~\cite{Lukezic_ICCV19_CDTB} with the same evaluation metrics. 
All the sequences are annotated on the basis of 5 attributes, 
namely, occlusion, dynamics change, motion change, size change, 
and camera motion.
RGB-T challenge constructs the dataset as a subset of RGBT234 with slight 
change in ground truth, which consists of 60 RGB-T public videos and 60 
sequestered videos. 
Compared with RGBT234, VOT-RGBT utilizes different evaluation metrics, 
i.e., EAO, to measure trackers. 
In VOT2019-RGBT, trackers need to be re-initialized, when tracking failure is detected~(the overlap between bounding box and ground truth is zero). 
Besides, VOT2020-RGBT introduces a new anchor mechanism to avoid a causal correlation 
between the first reset and the later ones~\cite{Kristan_ECCVW20_VOT20} instead of the re-initialization mechanism.
\vspace{-5mm}
\section{Experiments}\label{Sec:Experiment}
In this section, we conduct analysis on both public datasets and challenges 
from the overall comparison, attribute-based comparison, and speed. 
For fair comparison on speed, we refer to the device used (CPU or GPU), 
platform used (M: Matlab, MCN: Matconvnet, P: Python, and PT: PyTorch), 
and setting~(detailed information on CPU and GPU). \emph{The available codes and detailed description of trackers are collected and listed in the supplementary files.}

\subsection{Experimental Comparison on RGB-D Datasets}
\begin{table}\label{PTB_results}
	\tiny
	\caption{Experimental results on the PTB dataset. Numbers in parentheses indicate their ranks. 
	The top three results are in {\color{red}\bf red}, {\color{blue}blue}, and {\color{green}green} fonts.}
	\vspace{-5mm}
	\begin{center}  
		\begin{tabular}{|c|c|c|c|c|c|c|c|c|c|c|c|c|}
			\hline
			\multirow{2}{*}{Algorithm} & \multirow{2}{*}{\tabincell{c}{Avg.\\Rank}} & \multicolumn{3}{c|}{Target type} & \multicolumn{2}{c|}{Target size} & \multicolumn{2}{c|}{Movement} &\multicolumn{2}{c|}{Occlusion} & \multicolumn{2}{c|}{Motion type}\\
			\cline{3-13}
			& & Human & Animal & Rigid & Large & Small & Slow & Fast & Yes & No & Passive & Active\\
			\hline
			OTR~\cite{Kart_CVPR19_VSDCF}&2.91&\color{green}77.3(3)&68.3(6)&\color{green}81.3(3)&76.5(4)&\color{red} \bf 77.3(1)&\color{blue}81.2(2)&\color{red} \bf 75.3(1)&\color{blue}71.3(2)&84.7(6)&\color{red}\bf85.1(1)&\color{green}73.9(3)\\
			\hline
			WCO~\cite{Liu_Sensors20_WCO}&3.91&\color{blue}78.0(2)&67.0(7)&80.0(4)&76.0(5)&\color{blue}75.0(2)&78.0(7)&73.0(5)&66.0(6)&\color{blue}86.0(2)&\color{blue}85.0(2)&\color{red}\bf82.0(1)\\
			\hline
			TACF~\cite{Kuai_Sensors19_TACF}&4.64&76.9(5)&64.7(9)&79.5(5)&	\color{green}77.2(3)&74.0(4)&78.5(5)&\color{green}74.1(3)&68.3(5)&\color{green}85.1(3)&83.6(4)&72.3(5)\\
			\hline
			CA3DMS~\cite{Liu_TMM19_TAMS}&6&66.3(12)&\color{blue}74.3(2)&\color{red}\bf 82.0(1)&73.0(10)&\color{green}74.2(3)&\color{green}79.6(3)&71.4(8)&63.2(13)&\color{red} \bf88.1(1)&82.8(5)&70.3(8)\\
			\hline
			CSR-rgbd~\cite{Kart_ECCVW18_CSRDCF-rgbd}&6.36&76.6(6)&	65.2(8)&75.9(9)&75.4(6)&73.0(6)&79.6(3)&71.8(6)&\color{green}70.1(3)&79.4(10)&79.1(7)&72.1(6)\\
			\hline
			3DT~\cite{Bibi_CVPR16_ASR}&6.55&\color{red}\bf81.4(1)&64.2(11)&73.3(11)&\color{blue}79.9(2)&71.2(9)&75.1(12)&\color{blue}74.9(2)&\color{red} \bf72.5(1)&78.3(11)&79.0(8)&73.5(4)\\
			\hline
			DLST~\cite{An_ICPR16_DLS}&7&77.0(4)&69.0(5)&73.0(12)& \color{red} \bf 80.0(1)&70.0(12)&73.0(13)&74.0(4)&66.0(6)&85.0(5)&72.0(13)&\color{blue}75.0(2)\\
			\hline
			OAPF~\cite{Meshgi_CVIU16_OAPF}&7.27&64.2(13)&\color{red}\bf 84.8(1)&77.2(6)&72.7(11)&73.4(5)&\color{red}\bf 85.1(1)&68.4(12)&64.4(11)&\color{green}85.1(3)&77.7(10)&71.4(7)\\
			\hline
			CCF~\cite{Li_GSKI19_CCF}&7.55&69.7(10)&64.5(10)&\color{blue}81.4(2)&73.1(9)&72.9(7)&78.4(6)&70.8(9)&65.2(8)&83.7(7)&\color{green}84.4(3)&68.7(12)\\
			\hline
			OTOD~\cite{Xie_CIS19_OD}&8.91&72.0(8)&\color{green}71.0(3)&73.0(12)&74.0(7)&71.0(10)&76.0(9)&70.0(11)&65.0(9)&82.0(8)&77.0(12)&70.0(9)\\
			\hline
			DMKCF~\cite{Kart_ICPR18_DMDCF}&9&76.0(7)&58.0(13)&76.7(7)&72.4(12)&72.8(8)&75.2(11)&71.6(7)&69.1(4)&77.5(13)&82.5(6)&68.9(11)\\
			\hline
			DSKCF~\cite{Hannuna_RTIP19_DSKCF}&9.09&70.9(9)&70.8(4)&73.6(10)&73.9(8)&70.3(11)&76.2(8)&70.1(10)&64.9(10)&81.4(9)&77.4(11)&69.8(10)\\
			\hline
			DSOH~\cite{Camplani_BMVC2015_ASR}&11.45&67.0(11)&61.2(12)&76.4(8)&68.8(13)&69.7(13)&75.4(10)&66.9(13)&63.3(12)&77.6(12)&78.8(9)&65.7(13)\\
			\hline
			DOHR~\cite{Ding_FSKD15_OHR}&14&45.0(14)&49.0(14)&42.0(14)&48.0(14)&42.0(14)&50.0(14)&43.0(14)&38.0(14)&54.0(14)&54.0(14)&41.0(14)\\
			\hline
		\end{tabular}  
	\end{center}
	\vspace{-10mm}
\end{table}

{\flushleft \bf Overall Comparison}. PTB provides a website\footnote{ \href{http://tracking.cs.princeton.edu/}{http://tracking.cs.princeton.edu/}} 
for comprehensive evaluating RGB and RGB-D methods in an online manner. 
We collect the results of 14 RGB-D trackers on the website and sort them based on rank. 
The results are shown in Table~3. We list the Avg. Rank, SR and 
corresponding rank of each attribute. 
The Avg. Rank is calculated by averaging the rankings of all attributes.
According to Table~3, OTR achieves the best performance among all the competitors, 
which is based on the CF framework without deep features. 
Reason for the promising result is that the 3D construction provides a useful constraint for filter learning. 
The same conclusion can be obtained by CA3DMS and 3DT, which construct a 3D model to locate the target via 
mean-shift and sparse learning methods. 
These trackers with traditional features are competitive to the deep trackers. 
DL-based trackers~(WCO, TACF, and CSR-RGBD) achieve substantial performance, 
which indicates the discrimination of deep features.
CF-based trackers achieve various results and are the most widely-applied framework. 
Trackers based on original CF methods~(DMKCF, DSKCF and DSOH) perform significantly 
worse than those developed on improved CF~(OTR, WCO and TACF).
OTOD based on point cloud does not exploit the effectiveness of CNN and obtains the 10th rank on the PTB dataset.

{\flushleft \bf Attribute-based Comparison}. PTB provided 11 attributes from five aspects for comparison. 
CF-based trackers, including OTR, WCO, TACF, CSR-RGBD, and CCF, are not well-performed on tracking animals. 
As animals move fast and irregularly, these trackers with online learning are easy to drift. 
While the target is in small size, CF can provide precise tracking results. 
The occlusion handling mechanism contributes greatly to videos with target occlusion. 
The 3D mean shift method shows obvious advantage in tracking targets with rigid shape and no occlusion. 
OAPF obtains an above-average performance on tracking small objects, thereby indicating the effectiveness 
of the scale estimation strategy.

{\flushleft \bf Speed Analysis}. The speed report of RGB-D trackers are listed in Table~4. 
Most of the trackers cannot meet the requirements of real-time tracking. Trackers based on the improved CF framework~(OTR~\cite{Kart_CVPR19_VSDCF}, DMKCF~\cite{Kart_ICPR18_DMDCF}, CCF~\cite{Li_GSKI19_CCF}, 
WCO~\cite{Liu_Sensors20_WCO}, and TACF~\cite{Kuai_Sensors19_TACF}), are subject to their speed. 
Only two real-time trackers~(DSKCF~\cite{Hannuna_RTIP19_DSKCF} and DSOH~\cite{Camplani_BMVC2015_ASR}) are 
based on the original CF architecture.

\begin{table}\label{RGBD_speed}
	\scriptsize
	\vspace{-10mm}
	\caption{The speed analysis of RGB-D trackers.}
	\begin{center}  
		\begin{tabular}{|c|c|c|c|c|}
			\hline
			Trackers & Speed & Device & Platform & Setting \\
			\hline
			OTR~\cite{Kart_CVPR19_VSDCF}& 2.0 & CPU & Matlab & I7@3.6GHz \\
			\hline
			WCO~\cite{Liu_Sensors20_WCO}& 9.5 & GPU & M \& MCN & I7-6700@3.4GHz, GTX TITAN\\
			\hline
			TACF~\cite{Kuai_Sensors19_TACF}& 13.1 & GPU & M \& MCN & I7-6700K@4.0GHz \\
			\hline
			CA3DMS~\cite{Liu_TMM19_TAMS}& 63 & CPU & C++ & i7@3.6GHz\\
			\hline
			DLST~\cite{An_ICPR16_DLS}& 4.8 & CPU & M & I5-
			2400@3.10GHz\\
			\hline
			OAPF~\cite{Meshgi_CVIU16_OAPF}& 0.9 & CPU & M & -- \\
			\hline
			CCF~\cite{Li_GSKI19_CCF}& 6.3 & CPU & M & I7-6700@3.4GHz\\ 			
			\hline
			DMKCF~\cite{Kart_ICPR18_DMDCF}& 8.3 & CPU & M & I7@3.6GHz\\
			\hline
			DSKCF~\cite{Hannuna_RTIP19_DSKCF}& 40 & CPU & M \& C++ & I7-3770S@3.10GHz\\
			\hline
			DSOH~\cite{Camplani_BMVC2015_ASR}& 40 & CPU & M &I7-3770S@3.10GHz \\
			\hline
		\end{tabular}  
\end{center}
\vspace{-10mm}
\end{table}
\subsection{Experimental Comparison of RGB-T Datasets}
We select 14 trackers as our baseline to perform overall comparison of the GTOT and RGBT234 datasets. 
As only part of the trackers~(JMMAC, MANet, mfDiMP) release their codes, We run these trackers on these 
two datasets and record the performance of other trackers in their original papers. 
The overall results are shown in Table~5.

\begin{table}\label{RGBT_results}
	\scriptsize
	\caption{Experimental results on the GTOT and RGBT234 datasets.}
	\begin{center}  
		\begin{tabular}{|c|cc|cc|c|c|c|c|}
			\hline
			\multirow{2}{*}{Tracker} & \multicolumn{2}{c|}{GTOT} & \multicolumn{2}{c|}{RGBT234} & \multirow{2}{*}{Speed} &  \multirow{2}{*}{Device} &
			\multirow{2}{*}{Platform} & \multirow{2}{*}{Setting} \\
			\cline{2-5}
			& SR & PR & SR & PR & & & &\\
			\hline
			CMPP~\cite{Wang_CVPR20_CMPP} & \color{red}\bf 73.8 & \color{red}\bf 92.6 & \color{red}{\bf 57.5} & \color{red}{\bf 82.3} & 1.3 & GPU & PyTorch & RTX 2080Ti\\
			\hline
			JMMAC~\cite{Zhang_arxiv20_JMMAC} & \color{blue}73.2 & \color{blue} 90.1 & \color{blue}57.3 & 79.0 & 4.0 & GPU & MatConvNet & RTX 2080Ti\\
			\hline
			CAT~\cite{Li_ECCV20_CAT} & 71.7&88.9&\color{green}56.1 & \color{blue}80.4 & 20.0 & GPU & PyTorch & GTX 2080Ti\\
			\hline
			MaCNet~\cite{Zhang_Sensor20_MaCNet} & 71.4&88.0&\color{green}55.4 & 79.0 & 0.8 & GPU & PyTorch & GTX 1080Ti\\
			\hline
			TODA~\cite{Yang_ICIP19_TODA} & 67.7 & 84.3 & 54.5 & 78.7&0.3 & GPU & PyTorch & GTX1080Ti\\
			\hline
			DAFNet~\cite{Gao_ICCVW19_DAFNet}& 71.2&89.1&54.4&\color{green}79.6&23.0&GPU& PyTorch & RTX 2080Ti\\
			\hline
			MANet~\cite{Li_ICCVW19_MANet}& \color{green} 72.4&\color{green} 89.4 & 53.9 & 77.7 & 1.1 & GPU & PyTorch & GTX1080Ti\\
			\hline
			DAPNet~\cite{Zhu_MM19_DAPNet}& 70.7 & 88.2 & 53.7 & 76.6 & -- & GPU & PyTorch & GTX1080Ti\\
			\hline
			FANet~\cite{Zhu_arxiv18_FANet}& 69.8 & 88.5 & 53.2 & 76.4 & 1.3 & GPU & PyTorch & GTX1080Ti\\
			\hline
			CMR~\cite{Li_ECCV18_CMR} & 64.3 & 82.7 & 48.6 & 71.1 & 8.0 & CPU & C++ & -- \\
			\hline
			SGT~\cite{Li_MM17_RGBT210} & 62.8 & 85.1 & 47.2 & 72.0 & 5.0 & CPU & C++ & -- \\
			\hline
			mfDiMP~\cite{Zhang_ICCVW19_DIMP-RGBT} & 49.0 & 59.4 & 42.8 & 64.6 & 18.6 & GPU & PyTorch & RTX 2080Ti\\
			\hline 
			CSR~\cite{Li_TIP16_GTOT} & -- & -- & 32.8 & 46.3 & 1.6 & CPU & Matlab \& C++ & -- \\
			\hline
			L1-PF~\cite{Wu_ICIF11_L1-PF} &42.7&55.1&28.7&43.1&--&--&--&--\\
			\hline
			JSR~\cite{Liu_IS12_JSR} & 43.0 & 55.3 & 23.4 & 34.3 & 0.8 & CPU & Matlab & --\\ 
			\hline
		\end{tabular}  
\end{center}
\vspace{-10mm}
\end{table}		

{\flushleft \bf Overall Comparison}. All the high-performance trackers are equipped with the learned deep 
features and most of them are based on MDNet variants~(CMPP, MaCNet, TODA, DAFNet, MANet, DAPNet, and FANet), 
which achieve satisfactory results. 
The CF-based tracker~(JMMAC) obtains the second rank on GTOT and RGBT234 datasets, 
which combines appearance and motion cues. 
Compared with CF trackers, MDNet-based trackers can provide precise target position with higher PR, 
but are inferior to CF framework in scale estimation, reflecting on SR. 
Trackers with sparse learning technique~(CSR, SGT) are better than L1-PF based on the PF method. 
Although mfDiMP utilizes the state-of-the-art backbone, the performance is not positive. 
The main reason may be that mfDiMP utilizes different training data, which are generated by image 
translation methods~\cite{Zhang_TIP18_TIRtracking} and brings a gap between existing real RGB-T data.

\begin{figure}
	\centering
	\includegraphics[width=1.0\linewidth]{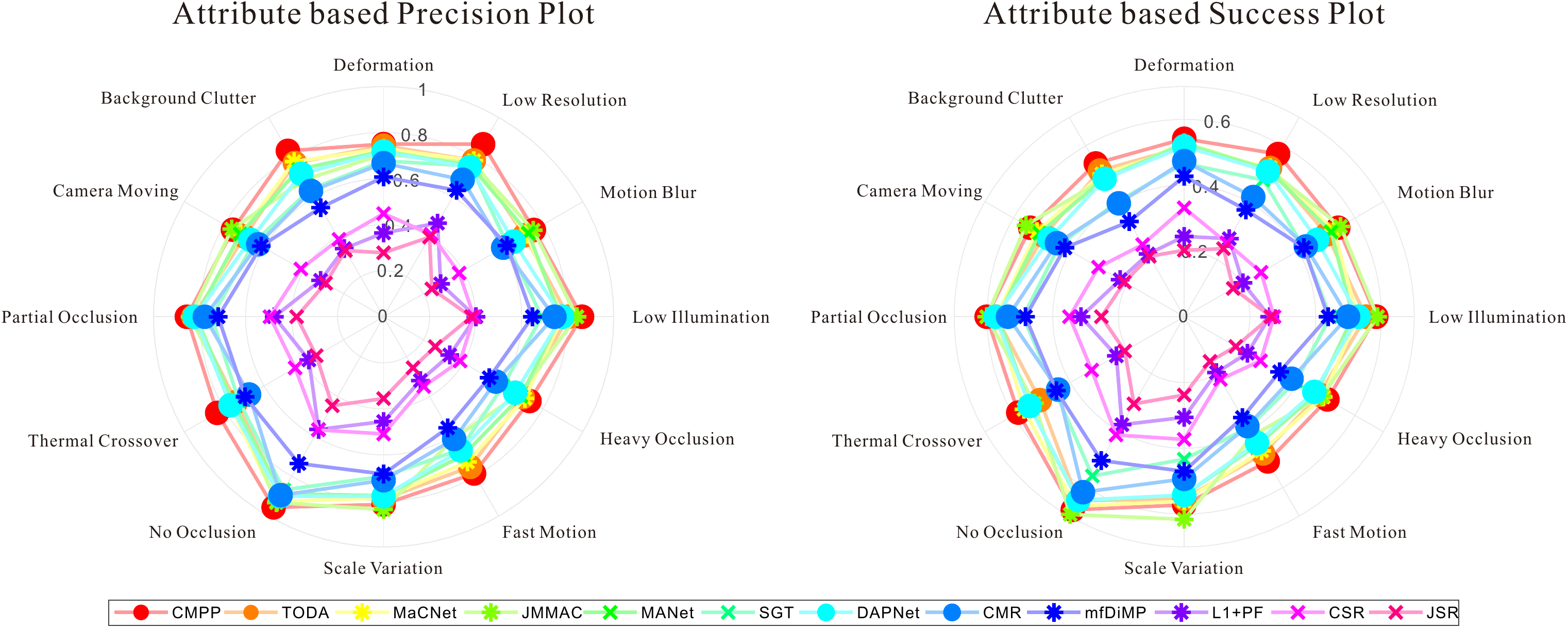}
	\vspace{-10mm}
	\caption{Attribute-based Comparison on RGBT234.}
	\label{fig:attr_RGBT234}
	\vspace{-5mm}
\end{figure}

{\flushleft \bf Attribute-based Comparison}. We conduct attribute-based comparison on RGBT234, 
as shown in Figure~\ref{fig:attr_RGBT234}. 
Improved MDNet-based trackers achieve satisfactory performance on low-resolution, deformation, 
background clutter, fast motion, and thermal crossover. 
Since modeling both camera motion and target motion, JMMAC has strength in camera movement and 
partial occlusion, but fails easily in fast-moving targets. 
This condition may result from CF-based trackers having a fixed search region. 
When the target moves outside the region, the target cannot be detected, thereby causing tracking failure. 
CMPP, which exploits the inter-modal and cross-modal correlation, have great promotion on low illumination, 
low resolution, and thermal crossover. 
Targets in these attributes have unavailable modality and CMPP can eliminate the gap between 
heterogeneous modalities. \emph{The detailed figure on attribute-based comparison can be found in supplementary file.}
 
{\flushleft \bf Speed Analysis}. For tracking speed, we list the platform and setting for fair comparison 
in Table~5. DAFNet based on a real-time MDNet variant achieves fast tracking with 23.0 fps. 
Although mfDiMP is equipped with ResNet-101, it achieves the second fastest tracker because most parts 
of the network are trained offline without tuning online. 
Other trackers are constrained by their low speed, which cannot be utilized in some real-time applications.

\subsection{Challenge Results on VOT2019-RGBD}
We list the challenge results in Table~6. 
Both original RGB trackers without utilizing depth information and RGB-D trackers are merged for evaluation. 
Trackers who obtain top three ranks on F-score, precision, and Recall, are designed with the same component 
and framework.
Unlike the PTB dataset, DL-based methods have potential performance on VOT-RGBD19, which results from 
these trackers utilizing large-scale visual datasets for offline training and equipping deeper networks. 
For instance, the original RGB tracker with DL framework also achieves excellent performance. 
Furthermore, occlusion handling is another necessary component of the high-performance tracker 
because VOT2019-RGBD focuses on long-term tracking with target frequent reappearance and 
out-of-view and most of the trackers are equipped with a redetection mechanism. 
The CF framework~(FuCoLoT, OTR, CSR-RGBD, and ECO) does not perform well, which may stem from the 
online updating using occlusion patches that degrade the discriminability of the model.

\begin{table}\label{VOT-RGBD19}
	\footnotesize
	\caption{Challenge results on VOT2019-RGBD dataset.}
	\begin{center}  
		\begin{tabular}{lcccccc}
			\hline
			\bf Tracker &\bf Rank & \bf Modality & \bf Type & \bf F-score & \bf Precison & \bf Recall\\
			\hline
			{SiamDW-D}~\cite{Zhang_CVPR19_SiamDW} & 1 & RGB-D &OR, DL  & \color{red}\bf 0.681 & \color{red}\bf 0.677 & \color{blue}0.685 \\
			{ATCAIS}~\cite{Kristan_ICCVW19_VOT19}  & 2 &RGB-D& OR, DL & \color{blue}0.676 & \color{green}0.643 & \color{red}\bf 0.712 \\
			{LTDSE-D}~\cite{Kristan_ICCVW19_VOT19}  & 3 & RGB-D & OR, DL & \color{green}0.658 & \color{blue}0.674 & \color{green}0.643 \\
			{SiamM-Ds}~\cite{Wang_CVPR19_SiamMask} & 4 & RGB-D & SE, DL & 0.455 & 0.463 & 0.447 \\
			{MDNet}~\cite{MDNet} & 5 & RGB & DL &0.455 & 0.463 & 0.447 \\
			{MBMD}~\cite{Kristan_ICCVW19_VOT19} & 6 & RGB & OR,DL & 0.441 & 0.454 & 0.429 \\
			{FuCoLoT}~\cite{Lukezic_ACCV18_FuCoLoT} & 7 &RGB &  OR, CF & 0.391 & 0.459 & 0.340 \\
			{OTR}~\cite{Kart_CVPR19_VSDCF}& 8 & RGB-D & CF, EF, Pre & 0.336 & 0.364 & 0.312 \\
			{SiamFC}~\cite{Bertinetto_ECCVW16_SiamFC} & 9 & RGB & DL & 0.333 & 0.356 & 0.312 \\
			{CSR-rgbd}~\cite{Kart_ECCVW18_CSRDCF-rgbd} & 10 & RGB-D & CF, EF, OR & 0.332 & 0.375 & 0.397 \\
			{ECO}~\cite{Danelljan_CVPR17_ECO} & 11 & RGB & CF & 0.329 & 0.317 & 0.342 \\
			{CA3DMS}~\cite{Liu_TMM19_TAMS} & 12 & RGB & MS, OR, Pre & 0.271 & 0.284 & 0.259\\
			\hline
		\end{tabular}  
	\end{center}
	\vspace{-5mm}
\end{table}	

\subsection{Challenge Results on VOT2019-RGBT}
For VOT2019-RGBT dataset shown in Table~7, JMMAC with exploiting both appearance and motion cues shows high accuracy and robust performance and obtains the highest EAO in a large margin. 
Early fusion is the primary manner of RGB-T fusion, while the late fusion method~(JMMAC) has 
great potential in improving tracking accuracy and robustness, which has not been fully utilized. 
All top six trackers are equipped with CNN as feature extractor, thereby indicating the powerful ability of CNN. 
SiamDW using a Siamese network is a general method that performs well in both RGB-D and RGB-T tasks. 
ATOM variants~(mfDiMP and MPAT) are used to handle RGB-T tracking.

\begin{table}\label{VOT-RGBT19}
	\footnotesize
	\caption{Challenge results on the VOT2019-RGBT dataset.}
	\begin{center}  
		\begin{tabular}{lcccccc}
			\hline
			\bf Tracker &\bf Rank & \bf Modality &\bf Type & \bf EAO & \bf Accuracy & \bf Robustness\\
			\hline
			JMMAC~\cite{Zhang_arxiv20_JMMAC} & 1 & RGBT & CF, LF & \color{red}\bf 0.4826 & \color{red}\bf 0.6649 & \color{red}\bf 0.8211\\
			SiamDW-T~\cite{Zhang_CVPR19_SiamDW} & 2 & RGBT & DL, EF & \color{blue} 0.3925 & 0.6158 & \color{green}0.7839\\
			mfDiMP~\cite{Zhang_ICCVW19_DIMP-RGBT} & 3 & RGBT & DL, EF & \color{green}0.3879 & 0.6019 & \color{blue} 0.8036\\
			FSRPN~\cite{Kristan_ICCVW19_VOT19} & 4 & RGBT & DL, EF & 0.3553 & \color{blue} 0.6362 & 0.7069\\
			MANet~\cite{Li_ICCVW19_MANet} & 5 & RGBT & DL, EF & 0.3463 & 0.5823 & 0.7010\\
 			MPAT~\cite{Kristan_ICCVW19_VOT19} & 6 & RGB &DL & 0.3180 & 0.5723 & 0.7242 \\
 			CISRDCF~\cite{Kristan_ICCVW19_VOT19} & 7 & RGBT & CF, EF & 0.2923 & 0.5215 & 0.6904 \\
 			GESBTT~\cite{Kristan_ICCVW19_VOT19}& 8 & RGBT & Lucas–Kanade & 0.2896 & \color{green}0.6163 & 0.6350 \\
			\hline 
		\end{tabular}  
	\end{center}  
	\vspace{-7mm}
\end{table}	
\vspace{-5mm}
\section{Further Prospects}\label{Sec:Discussion}
\subsection{Model Design}
{\flushleft \bf Multi-modal fusion}. Compared with tracking with unimodality data, multi-modal tracking 
can easily exploit a powerful data fusion mechanism. 
Existing methods mainly focus on feature fusion, whereas the effectiveness of other fusion types has  
not been exploited. 
Compared with early fusion, late fusion eliminates the bias that heterogeneous features may be learned 
from different modalities. 
Furthermore, another advantage of late fusion is that we can utilize various methods to model each 
modality independently. 
The hybrid fusion method combining the early and late fusion strategies has been used in image segmentation~\cite{Bendjebbour_TGRS01_HF1} and sports video analysis~\cite{Xu_TMCCA06_HF2}, 
which is also a better choice for multi-modal tracking.

{\flushleft \bf Specific Network for Auxiliary Modality}. As the gap of different modalities exists 
and the semantic information is also heterogeneous, traditional methods use different features 
to extract more useful data~\cite{Lan_PRL18_MCASR,Liu_IS12_JSR,Garcia_JDOS12_AMC}. 
Although sufficient works on network structures for visible image analysis have been conducted, 
the specific architecture for depth and thermal maps has not been deeply explored. 
Thus, DL-based methods~\cite{Li_ICCVW19_MANet,Gao_ICCVW19_DAFNet,Zhu_MM19_DAPNet,Zhang_ICCVW19_DIMP-RGBT} 
trade the data in auxiliary modality as an additional dimension of the RGB image with the same network architecture, 
e.g., VGGNet and ResNet, and extract the feature in the same level~(layer). 
A crucial task is to design a network for processing multi-modal data. Since 2017, AutoML method,
especially neural architecture search~(NAS), has been popular which design the architecture automatically 
and obtain highly competitive results in many areas, such as image classification~\cite{Liu_ICLR19_DARTS}, and recognition~\cite{Zoph_CVPR18_NASNet}. 
However, researchers pay less attention to NAS method for multi-modal tracking, which is a good direction to explore. 

{\flushleft \bf Multi-modal Tracking with Real-time Speed}. The additional modality multiplies the computation, 
which causes difficulty for the existing tracking frameworks to achieve the requirements of real-time performance. 
A speed-up mechanism needs to be designed, such as feature selection~\cite{Zhu_MM19_DAPNet}, 
knowledge distillation technology, and others. 
Furthermore, Huang \emph{et al.}~\cite{Huang_ICCV17_EAST} propose a trade-off method, where the agent 
decides which layer is more suitable for accurate localization, thereby providing 100 times speed boost.

\subsection{Dataset Construction}
{\flushleft \bf Large-scale Dataset for Training}. With the emergence of deep neural networks, 
more powerful methods are equipped with CNN to achieve accurate and robust performance. 
However, the existing datasets focus on testing with no training subset. 
For instance, most DL-based trackers use the GTOT dataset as the training set when testing RGBT234, 
which has a small amount of data with limited scenes. 
The effectiveness of DL-based methods has not been fully exploited. 
Zhang \emph{et al.}~\cite{Zhang_ICCVW19_DIMP-RGBT} generate synthetic thermal data from the numerous 
existing visible datasets by using the image translation method~\cite{Zhang_TIP18_TIRtracking}. 
However, this data augmentation does not bring significant performance improvement. 
Above all, constructing a large-scale dataset for training is the main direction for multi-modal tracking.

\begin{figure}[t]
	\centering
	\includegraphics[width=1.0\linewidth]{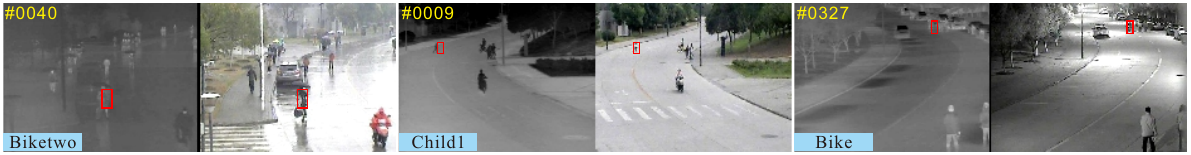}
	\vspace{-8mm}
	\caption{Unregistration examples in RGBT234 dataset. We show the ground truth of visible modality 
	in both images. The coarse bounding box degrades the discriminability of the model.}
	\label{fig:unaligned_images}
	\vspace{-5mm}
\end{figure}

{\flushleft \bf Modality Registration}. As multi-modal data is captured by different sensors and 
the binocular camera has a parallax error that cannot be ignored when the target is small and 
has low resolution, registering the data in spacial and temporal aspects is essential. 
As shown in Figure~\ref{fig:unaligned_images}, the target is out of the box and the model is 
degraded by learning meaningless background information. 
In the VOT-RGBT challenge, the dataset ensures the precise annotation in infrared modality 
and the misalignment of the RGB image is required to be handled by the tracker. 
We state that the image pre-registration process is necessary during dataset construction 
by cropping the shared visual field and applying image registration method.

{\flushleft \bf Metrics for Robustness Evaluation}. In some extreme scenes and weather conditions, 
such as rainy, low illumination and hot sunny days, visible or thermal sensors cannot provide meaningful data. 
The depth camera cannot obtain precise distance estimation when the object is far from the sensor. 
Therefore, a robust tracker needs to avoid tracking failure when any of the modality data is unavailable 
during a certain period. 
To handle this case, both complementary and discriminative features have to be applied in localization. 
However, none of the datasets measures the tracking robustness with missing data. 
Thus, a new evaluation metric for tracking robustness needs to be considered.
\vspace{-5mm}
\section{Conclusion}
In this study, we provide an in-depth review of multi-modal tracking.
First, we conclude multi-modal trackers in a unified framework, 
and analyze them from different perspectives, including auxiliary 
modality purpose and tracking framework.
Then, we present a detailed introduction on the datasets for multi-modal 
tracking and corresponding metrics.
Furthermore, a comprehensive comparison of five popular datasets is 
conducted and the effectiveness of trackers belonging to various types 
are analyzed in the views of overall performance, attribute-based performance,
and speed.
Finally, as an emerging field, several possible directions are identified to
facilitate the improvement of multi-modal tracking.
The comparison results and analysis will be available at \href{https://github.com/zhang-pengyu/Multimodal_tracking_survey}{https://github.com/zhang-pengyu/Multimodal\_tracking\_survey}.

\bibliography{mybibfile}
\end{document}